\documentclass[conference]{IEEEtran}
\IEEEoverridecommandlockouts
\usepackage{cite}
\usepackage{amsmath,amssymb,amsfonts}
\usepackage{algorithmic}
\usepackage{graphicx}
\usepackage{textcomp}
\usepackage{xcolor}
\usepackage{float}
\usepackage{hyperref}
\def\BibTeX{{\rm B\kern-.05em{\sc i\kern-.025em b}\kern-.08em
    T\kern-.1667em\lower.7ex\hbox{E}\kern-.125emX}}
\begin{document}

\title{Automated Interpretation of Non-Destructive Evaluation Contour Maps Using Large Language Models for Bridge Condition Assessment\\
}

\author{\IEEEauthorblockN{ Viraj Nishesh Darji}
\IEEEauthorblockA{\textit{School of Computing} \\
\textit{George Mason University}\\
Fairfax, VA, USA \\
vdarji@gmu.edu}
\and
\IEEEauthorblockN{Callie C. Liao}
\IEEEauthorblockA{\textit{College of Science} \\
\textit{George Mason University}\\
Fairfax, VA, USA \\
cliao3@gmu.edu}
\and
\IEEEauthorblockN{Duoduo Liao}
\IEEEauthorblockA{\textit{School of Computing} \\
\textit{George Mason University}\\
Fairfax, VA, USA \\
dliao2@gmu.edu}
}

\maketitle

\begin{abstract}
Bridge maintenance and safety are essential for transportation authorities, and Non-Destructive Evaluation (NDE) techniques are critical to assessing structural integrity. However, interpreting NDE data can be time-consuming and requires expertise, potentially delaying decision-making. Recent advancements in Large Language Models (LLMs) offer new ways to automate and improve this analysis. This pilot study introduces a holistic assessment of LLM capabilities for interpreting NDE contour maps and demonstrates the effectiveness of LLMs in providing detailed bridge condition analyses. It establishes a framework for integrating LLMs into bridge inspection workflows, indicating that LLM-assisted analysis can enhance efficiency without compromising accuracy. In this study, several state-of-the-art LLMs are explored with prompts specifically designed to enhance the quality of image descriptions, which are applied to interpret five different NDE contour maps obtained through technologies such as Ground Penetrating Radar (GPR), Electrical Resistivity (ER), Impact-Echo (IE), and Ultrasonic Surface Waves (USW) for assessing bridge conditions. Each LLM model is evaluated based on its ability to produce detailed descriptions, identify defects, provide actionable recommendations, and demonstrate overall accuracy. The research indicates that the ChatGPT-4, Claude 3.5 Sonnet, CogVLM2, and ShareGPT4V models provide better image descriptions, effectively covering a wide range of topics related to the bridge's condition. The outputs from these four models are subsequently summarized using five different LLMs to form a comprehensive overview of the bridge. Notably, LLMs ChatGPT-4 and Claude 3.5 Sonnet generate more detailed and effective summaries. The findings suggest that LLMs have the potential to significantly improve the efficiency and accuracy of infrastructure evaluation processes. This pilot study presents an innovative approach that leverages LLMs for image captioning in parallel and summarization, enabling faster decision-making in bridge maintenance and enhancing infrastructure management and safety assessments.

\end{abstract}


\section{Introduction}
The maintenance and safety of bridges are of high importance for transportation authorities worldwide. Non-Destructive Evaluation (NDE) techniques play a crucial role in identifying bridge conditions and providing valuable insights into the structural integrity of the bridge. However, interpreting NDE data often requires expertise and can take a significant amount of time which can delay the decision-making process. Recent advancements in Large Language Models (LLMs) have opened new ways of automating and enhancing the analysis of technical data. LLMs have demonstrated remarkable capabilities in understanding and describing complex visual information from various fields. This research explores the potential of leveraging LLMs to explain NDE contour maps and identify bridge conditions, intending to provide more accessible insights to engineers and decision-makers efficiently. 

The primary objective of this pilot study is to analyze NDE data using LLMs and evaluate the performance of the LLMs based on the accuracy and quality of the data. A diverse set of LLMs, including both open-source and proprietary solutions, are employed to achieve the objective. 

The critical novelty of this research is multifold. First, a holistic assessment is provided for gaining access to a wide range of LLM capabilities in identifying the data of NDE contour maps. Second, the effectiveness of multiple LLMs in giving detailed interpretations of bridge conditions is demonstrated with a novel approach that uses secondary LLM analysis to synthesize and consolidate the findings. Third, we have developed a practical framework for integrating LLMs into bridge inspection workflows. Our findings indicate that LLM-assisted analysis can serve as a valuable support tool for existing bridge inspection reports, potentially reducing the time required for data interpretation without compromising accuracy. This study opens new avenues for more efficient bridge maintenance decision-making, contributing to the broader fields of infrastructure management and safety assessment research. 

\section{Related Work}

Recent breakthroughs in deep learning have contributed significantly to automated infrastructure inspection and evaluation processes. The existing work can be broadly categorized into two key areas: a) bridge inspection systems using image captioning techniques, and b) attention mechanisms and captioning architectures for technical image interpretations.

\subsection{Attention Mechanisms and Technical Image Captioning}
The Transformer model, as first proposed by Vaswani et al. \cite{b3}, is the backbone for many of today's state-of-the-art image interpretation systems. They demonstrated that self-attention mechanisms alone can provide state-of-the-art performance for sequence transduction tasks without recurrence or convolution. The work was targeted at text translation; however, the principles they laid down have later become very important in image-to-text applications, including our approach to NDE contour map interpretation.
Al-Malla et al. \cite{b4} further enhanced this by adding object detection features and an attention mechanism to the image captioning model. Their work focused on natural image captioning, but the multi-modal feature extraction techniques they employed provide a valuable foundation for our application in technical NDE image interpretation.

\subsection{Bridge Inspection Systems Using Image Captioning}
Bridge inspection systems using image captioning leverage deep learning, particularly Convolutional Neural Networks (CNNs), to analyze bridge images and generate descriptive captions of visible defects like cracks and corrosion. These systems improve inspection efficiency and accessibility, especially for less experienced engineers. However, these approaches are limited to surface-level damage detection in photographs and do not address more complex data sources such as NDE contour maps.
Li et al. \cite{b1} proposed BDCD-Net, an end-to-end image captioning framework that is specifically developed for the description of bridge damage, incorporating an adaptive attention mechanism to provide a comprehensive description of the bridge damage, including component types, damage categories, and spatial information, while demonstrating the feasibility of automatic bridge inspection. However, it mainly focuses on visible damage in photographic images, rather than interpreting the contour maps from NDE data, as in our approach.
Chun et al. \cite{b2} developed a web-based bridge inspection system that generates explanatory texts from bridge photographs and enables continuous learning through user interactions, aiming to support less experienced engineers and staff. While their work focuses on visual inspection, our approach extends to subsurface analysis via NDE contour maps and employs multiple large language models for more comprehensive interpretation.

\quad

Although these works represent important advances in the active areas of image captioning and bridge inspection, the critical link in the automatic interpretation of NDE contour maps remains missing. Whereas past research was limited to visible damage in photographic images, our work applies these principles to the more specialized domain of NDE data visualization. By building on the latest improvements of LLMs in this research and further adapting existing techniques for captioning to tackle the special difficulties of interpreting technical NDE contour maps, the results would be more accessible to a wider range of stakeholders in the bridge maintenance process.

\section{Methodology}
\begin{figure}[htbp]
\centerline{\includegraphics[width=1.0\columnwidth]
{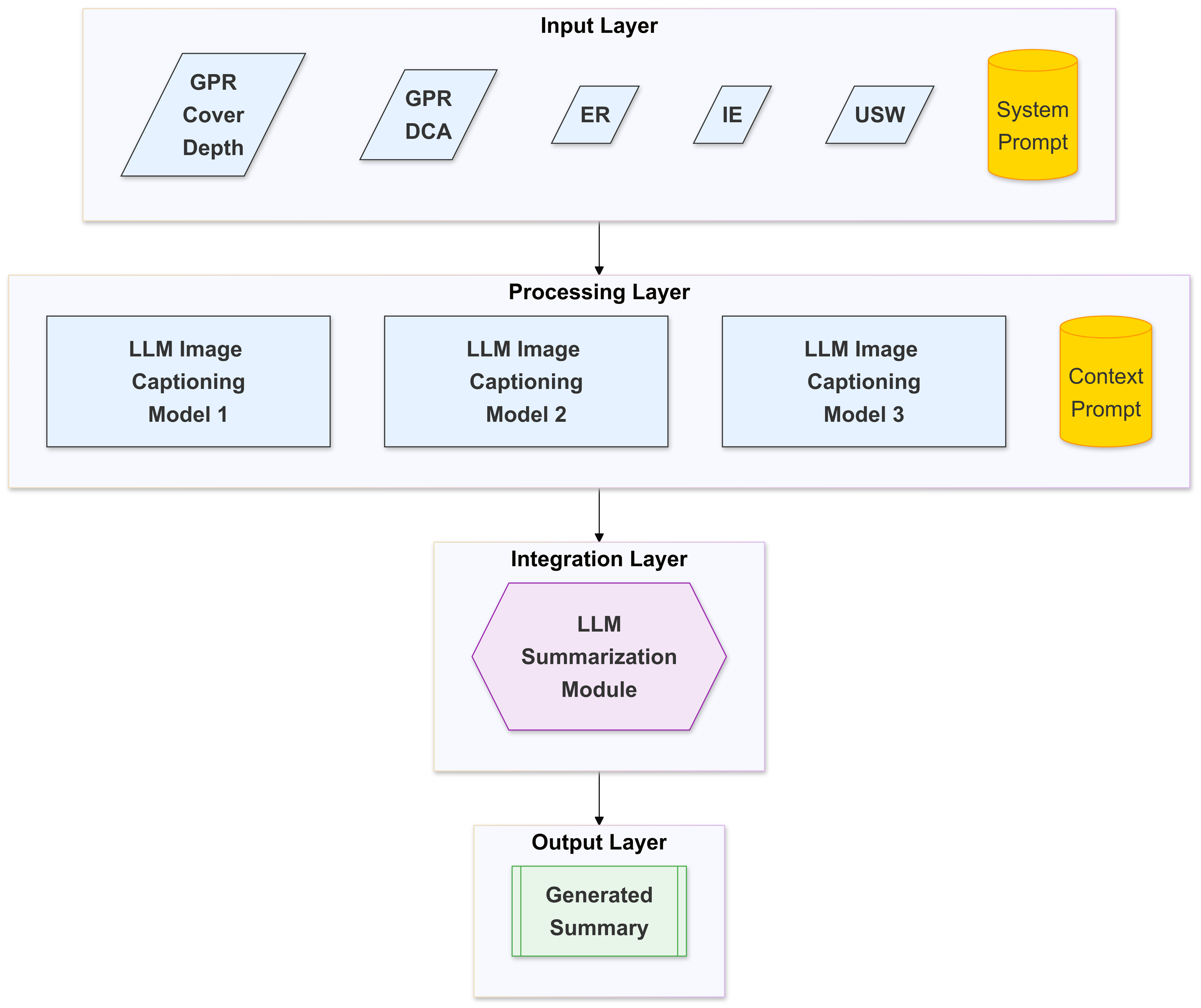}}
\caption{The framework of automated interpretation of NDE contour maps Using LLMs for bridge condition assessment}
\label{fig1}
\end{figure}
This section describes the structured methodology for interpreting LLMs to leverage NDE contour map data assessing bridge conditions. The process falls into three stages as shown in Fig.~\ref{fig1}: initial processing of data, multi-model image captioning, and summarization analysis.
\subsection{Data Input and Preparation}
The process inputs five different NDE contour map images, each representing different NDE measurement technologies used to scan the bridge, including Ground Penetrating Radar (GPR), Electrical Resistivity (ER), Impact Echo (IE), and Ultrasonic Surface Waves (USW). These maps provide variable information about the bridge's structure condition, with each image captured using different measurement parameters. A well-framed initial prompt is designed to instruct the LLMs in interpreting these technical visualizations. The prompt ensures consistency across the models in terms of extracting relevant technical information.

\subsection{Multi-Model Image Captioning}
The second step involves NDE contour maps for various parallel image captioning through LLMs. This parallel processing approach facilitates cross-referencing of findings across different models, leading to more reliable data interpretation. Image captioning is driven by two sets of prompts: the first one requests the extraction of relevant technical features from the NDE contour maps, while the second refines the interpretation based on specific requirements and technical parameters. This prompt structure ensures that the models focus on the most relevant aspects of the contour maps while maintaining technical accuracy in the interpretations.
\subsection{Summarization Analysis}

This stage involves consolidated analysis using LLM summarization, which synthesizes outputs from multiple image captioning models in their different interpretations toward one coherent analysis. The summarization model integrates the interpretations from different models' outputs, prioritizing critical information about the bridge condition and generating actionable recommendations based on consolidated findings.

\subsection{Output Generation}
The methodology culminates in the generation of a comprehensive output that encompasses several key aspects, including structural integrity assessment, defect identification and classification, and maintenance recommendations. 

\quad

This approach offers the advantage of leveraging multiple LLMs while maintaining a more structured and systematic approach to interpreting technical data. The multi-stage process enhances the robustness of the analysis and makes complex NDE data more accessible to the stakeholders involved in bridge maintenance and decision-making. By integrating the strengths of multiple models with structured summarization, this system is capable of providing a comprehensive analysis of NDE data that is actionable, thereby supporting effective and informed decision-making on bridge maintenance.

\section{Experimental Results and Data Analysis}
\subsection{Data Collection}
The data for this study are sourced from the Federal Highway Administration (FHWA) bridge database. As of July 23rd, 2024, 38 bridges in that database included NDE data. For this pilot study, we looked at only one bridge located in Mississippi, from which we extracted 5 unique NDE contour map datasets as shown in Fig. 2-6 in Appendix~\ref{NDE Contour Maps}. The selected structure has Structure Number 11002200250051B and LTBP Bridge Number 28-000008\cite{b5}.

Contour maps used in this study are representations of five different NDE measurement technologies and parameters. First, GPR measurements provided two different datasets: cover depth in inches (Fig.~\ref{fig2}) and Depth-Corrected Attenuation (DCA) at the top rebar level in decibels (Fig.~\ref{fig3}). These GPR measurements provide valuable information about both the internal structure of the bridge deck and possible trends in deterioration. Second, ER measurements are collected as an indicator of concrete properties and corrosion activity (Fig.~\ref{fig4}). IE tests in kHz (Fig.~\ref{fig5}) were applied to evaluate the integrity of the concrete elements and their eventual delamination. Finally, the measurements of USW, given in ksi (Fig.~\ref{fig6}), provided information about the modulus of the concrete, hence informing about its mechanical properties.
A range of measurement techniques have been selected to provide a comprehensive assessment of the bridge's structural condition, utilizing various physical parameters. Each measurement method offers unique insights into different aspects of the structure's health, and when analyzed collectively, they enable a more thorough and complete evaluation.

\subsection{LLMs for Image Captioning  }
The presented models in this paper represent state-of-the-art multimodal LLMs integrated with image captioning capabilities, each bearing different functionalities. The vit-gpt2-image-captioning model combines Vision Transformers with GPT-2 for generating image captions, therefore placing a heavy emphasis on the interaction between visual and textual data\cite{b6}. BLIP's large image captioning model integrates vision and language, leveraging the capabilities of the Vision Transformer and BERT to deliver highly accurate captions\cite{b7}. Paligemma\cite{b8} and its enhanced version, Paligemma rich captions\cite{b9}, are based on a pre-trained model for image captioning, fine-tuned to obtain longer descriptions. Florence-2 is a multimodal model that allows both image captioning and cross-modal understanding, making it effective for detailed visual understanding\cite{b10}. ShareGPT4V, a model based on LLaMA, can provide highly detailed captions and explanations; it focuses on delivering useful information extracted from images\cite{b11}. CogVLM2 is a state-of-the-art, next-generation advanced vision-language model designed for high-quality, in-depth captioning and analysis of visual content\cite{b12}. ChatGPT-4 is a powerful multimodal LLM that processes both textual and visual inputs to generate high-quality descriptive outputs\cite{b15}. Claude 3.5 Sonnet is a general multimodal LLM, performing proficiently in interpreting and captioning images into clear, accurate descriptive texts\cite{b16}. 

\subsection{Evaluation of Image Captioning Performance in LLMs }
In this study, these models have been evaluated based on their ability to generate detailed captions and their performance in analyzing the visual data against the NDE contour maps. This testing included an evaluation phase where multiple prompt variations were tested with the models to optimize their interpretation capabilities. While one model, vit-gpt2-image-captioning was used without any prompts,  another model, the BLIP image captioning large model, had been previously tested both with and without any prompts. In contrast, all other models underwent extensive optimization of prompts to improve their performance and interpretation accuracy. Prompt optimization played a crucial role in enhancing the quality of the output from LLMs in analyzing NDE data. The original prompt was intended to provide a simple description, but it underwent several iterative refinement steps to ultimately produce complete and relevant responses. Finally, the tone and content of the prompt were carefully crafted as shown in Appendix~\ref{prompt} to elicit a professional stance from the model, emphasizing the key visual features and providing a thorough assessment of the issue at hand. It underscores the importance of clear, contextual, and role-assigned prompts in enhancing the analysis capability of LLMs.

The models were evaluated based on a structured framework consisting of four key metrics: \textit{Relevance}, \textit{Usefulness}, \textit{Coverage}, and \textit{Specificity}. \textit{Relevance} refers to accuracy in the description of the pictures, specifically assessing whether the output captures most of the key features and the bridge condition. \textit{Usefulness} considers its applicability for both the engineer and the decision-maker. \textit{Coverage} analyzes whether the captions cover various aspects of the images, including all potential defects and overall conditions. \textit{Specificity} examines the level of detail provided, assessing the inclusions of subtle observations.

These ratings were developed for this pilot study to ensure consistency and alignment with its objectives. The evaluation consists of a qualitative analysis of output from each model, relative to pre-developed criteria for each metric, with assessments conducted by researchers and one domain expert. Binary assessments (Yes/No) are given if a model consistently upholds the standard in most test cases. The overall marks, ranging from 1 to 5, are combined to reach an overall score for each model's performance. A score of 5 would correspond with the assessment "Yes" in all four categories; on the contrary, a score of 1 would correspond with the assessment "No" in all categories. Table~\ref{tab1} presents the comparison of the performance of the models.

\begin{table}[htbp]
\caption{Image Captioning LLM Comparison}
\begin{center}
\setlength{\tabcolsep}{2.5pt} 
\begin{tabular}{|c|c|c|c|c|c|}
\hline
\textbf{LLM} & \textbf{Relevant}& \textbf{Usability}& \textbf{Coverage}& \textbf{Specificity}& \parbox[c]{1.1cm}{\centering \textbf{Overall} \\ \textbf{Rating}} \\
\hline
Claude 3.5 Sonnet& Yes& Yes& Yes& Yes& 5 \\
ChatGPT-4& Yes& Yes& Yes& Yes& 5 \\
CogVLM2& Yes& Yes& Yes& Yes& 5 \\
ShareGPT4V& Yes& Yes& Yes& No& 4 \\
Florence-2& Yes& No& Yes& No& 3 \\
Paligemma FT$^{\mathrm{*}}$& Yes& No& Yes& No& 3 \\
Paligemma& Yes& No& No& No& 2 \\
BLIP large& No& No& No& No& 1 \\
vit-gpt2& No& No& No& No& 1 \\
\hline
\multicolumn{3}{l}{$^{\mathrm{*}}$FT is Finetuned}
\end{tabular}
\label{tab1}
\end{center}
\end{table}

From the detailed evaluation, four models, ShareGPT4V, CogVLM2, ChatGPT-4, and Claude 3.5 Sonnet, showed superior performances, with overall ratings of 4 and above.  A detailed illustration of the capabilities of the Claude 3.5 Sonnet models is demonstrated in Appendix~\ref{imagecaptioning} using the prompt shown in Appendix~\ref{prompt}.

\subsection{Summarization for LLMs and Evaluation}
Outputs from these four top-performing image captioning models in this study were subjected to further summarization by five LLMs: ChatGPT-4\cite{b15}, Claude 3.5 Sonnet\cite{b16}, Mistral\cite{b13}, Gemini\cite{b17}, and Llama3\cite{b14}. This extra layer of analysis was aimed at consolidating the various interpretations into cohesive, actionable insight about the status/condition of the bridges.

These summarization models were evaluated using a structured framework based on three major metrics: \textit{Completeness}, \textit{In-Depth Coverage}, and \textit{Formatting \& Presentation}. \textit{Completeness} described the degree to which the output contained all the information that would make the summary comprehensive enough to hit the main points. Similarly, \textit{In-depth Coverage} checks for technical correctness and depth of analysis, whether the knowledge given was interesting and informative, and the satisfaction of requirements for professional review, whereas \textit{Formatting \& Presentation} checks for clarity, logical flow, as well as the format of output, thereby showing how well the summaries will be able to convey information in a clear and comprehensible way.

These scores were developed specifically for this pilot study, which helps standardize this research and meet the specific objectives of the study. Each model was matched against these criteria and rated on a scale from 1 to 5, with evaluations conducted by researchers and one domain expert. A score of 5 indicates exceptional performance, while 1 represents significant deficiencies in meeting the evaluation criteria. Table~\ref{tab2} provides a comparison of these summarization models.

\begin{table}[htbp]
\caption{Summarization LLM Comparison}
\begin{center}
\setlength{\tabcolsep}{2.5pt} 
\begin{tabular}{|c|c|c|c|c|}
\hline
\textbf{LLM} & \textbf{Completeness}& \parbox[c]{1.1cm}{\centering \textbf{In-depth} \\ \textbf{Coverage}}& \parbox[c]{1.9cm}{\centering \textbf{Formatting \&} \\ \textbf{Presentation}}& \parbox[c]{1.1cm}{\centering \textbf{Overall} \\ \textbf{Rating}} \\
\hline
ChatGPT-4& 5& 5& 5& 5 \\
Claude 3.5 Sonnet& 5& 4& 5& 4.67 \\
Gemini& 3& 3& 4 &3.33 \\
Mistral& 4& 3& 4& 3.67 \\
Llama3& 3& 3& 4& 3.33 \\
\hline
\end{tabular}
\label{tab2}
\end{center}
\end{table}

Among these, ChatGPT-4 ranked the highest with a perfect overall rating of 5.00. The second highest was Claude 3.5 Sonnet, having an overall rating of 4.67ma3 performed relatively poorly with an overall rating ranging from 3.33 to 3.67.
A typical output of the ChatGPT-4 summarization is shown in the Appendix \ref{summairzation}, where it systematically brought together many model interpretations into one coherent analysis of the bridge conditions.

\section{Conclusion}
This pilot study has demonstrated the effectiveness of LLMs in understanding the NDE contour map data of bridge inspection processes. Overall, our assessment of nine image captioning models and five summarization models indicated that ChatGPT-4, CogVLM2, Claude 3.5 Sonnet, and ShareGPT4V were exceptionally good at interpreting even complex NDE contour data interpretation, while ChatGPT-4 and Claude 3.5 Sonnet were very capable with summaries inclusive of comprehensive details.
These results have validated the potential of LLMs in three key areas: interpretation of the bridge condition, analysis of NDE contour maps, and generation of maintenance recommendations. This breakthrough has significant implications for accelerating infrastructure management bridge inspection processes, with complex NDE data made more accessible to a wide range of stakeholders. The system's capability for rapid and comprehensible interpretations results in more time- and resource-effective decision-making within bridge maintenance workflows. However, even as these models demonstrate remarkable abilities, they can only complement human knowledge, expertise, and judgment, and the periodic inconsistencies in the outputs underscore the point of validation by experts.

\newpage 

\appendix
\subsection{NDE Contour Maps}
\label{NDE Contour Maps}
\begin{figure}[h]
\centerline{\includegraphics[width=1.0\columnwidth]
{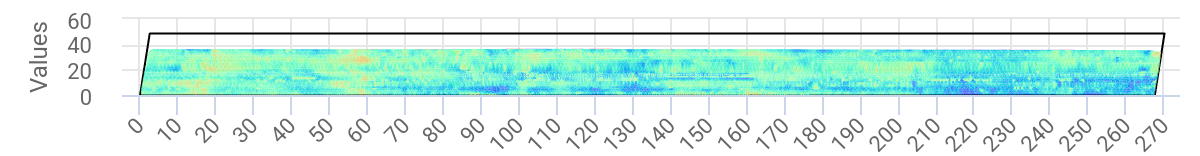}}
\caption{Cover Depth (inch) based on GPR Measurements \cite{b5}}
\label{fig2}
\centerline{\includegraphics[width=1.0\columnwidth]
{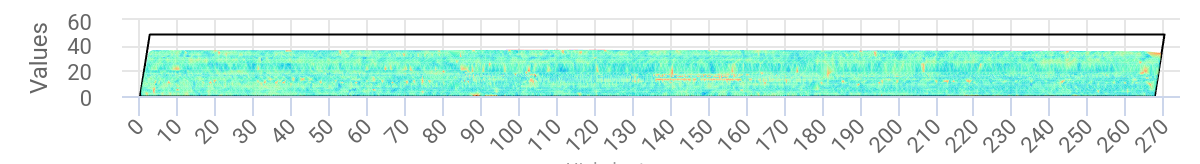}}
\caption{DCA At Top Rebar Level (dB) based on GPR Measurements \cite{b5}}
\label{fig3}
\centerline{\includegraphics[width=1.0\columnwidth]{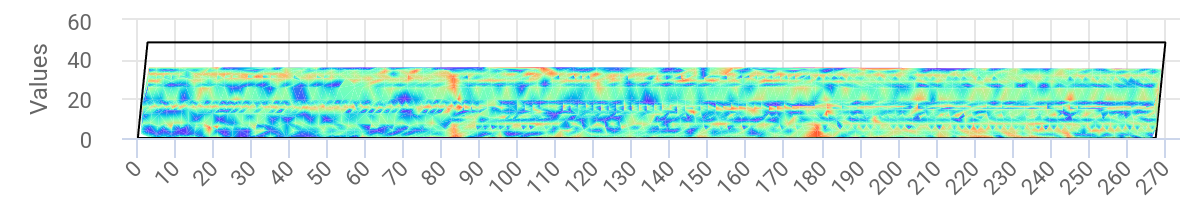}}
\caption{Electrical Resistivity \cite{b5}}
\label{fig4}
\centerline{\includegraphics[width=1.0\columnwidth]{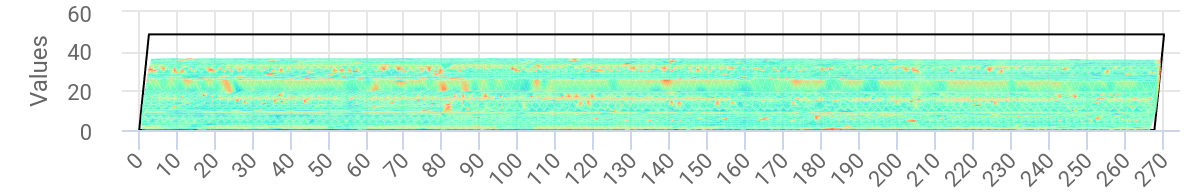}}
\caption{Frequency (kHz) based on IE Measurements \cite{b5}}
\label{fig5}
\centerline{\includegraphics[width=1.0\columnwidth]{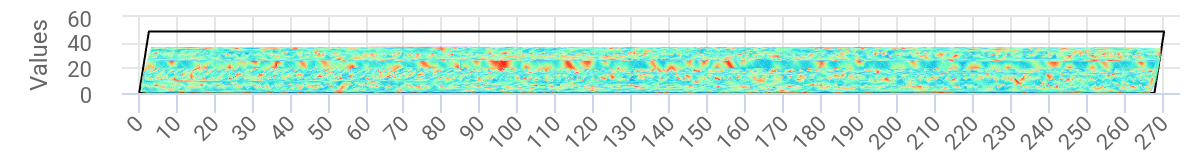}}
\caption{Modulus (ksi) based on USW Measurements \cite{b5}}
\label{fig6}
\end{figure}

\subsection{Prompt for Image Captioning}
\label{prompt}
\begin{table}[htbp]
\caption{Prompt for Image Captioning}
\begin{center}
\begin{tabular}{|p{8cm}|} 
\hline
\\
\textbf{Prompt}: You are a structural engineer analyzing a bridge. You have used different NDE technologies that provided outputs as shown in these 5 images; the x- and y-axes of the graphs are in feet. Analyze these images and describe the condition of the bridge, determine if it needs any repairs, and identify which areas of the bridge are in poor condition.
\\
\\ \hline

\end{tabular}
\label{tab3}
\end{center}
\end{table}

\subsection{Image Captioning by Claude 3.5 Sonnet}
\label{imagecaptioning}
\begin{table}[H]
\caption{Image Captioning by Claude 3.5 Sonnet\cite{b16}}
\begin{center}
\begin{tabular}{|p{8cm}|} 
\hline
\\
As a structural engineer analyzing these non-destructive evaluation (NDE) results, I can provide the following assessment of the bridge condition:
    \begin{enumerate}
        \item \textbf{Impact-Echo (IE) Measurements (Fig. \ref{fig2}):} The frequency distribution appears relatively uniform across the bridge length, with most areas showing frequencies in the mid-range (green). However, there are some spots with lower frequencies (darker blue), particularly near the 80-100 ft and 220-240 ft marks. Lower frequencies can indicate potential delamination or internal defects.
        \item \textbf{Ultrasonic Surface Waves (USW) Measurements (Fig. \ref{fig3}):} The modulus values vary across the bridge, with some areas showing lower values (blue) and a few spots with very low values (dark blue/purple). These areas, particularly around 90-100 ft and 230-250 ft, may have reduced concrete quality or strength.
        \item \textbf{Ground Penetrating Radar (GPR) Measurements (Fig. \ref{fig4}):} The attenuation levels are generally consistent, but there are areas with higher attenuation (darker blue), especially in the first 50 ft and around the 230-250 ft mark. Higher attenuation could indicate increased moisture content or deterioration.
        \item \textbf{Electrical Resistivity Measurements (Fig. \ref{fig5}):} There's significant variation in resistivity across the bridge. Areas with lower resistivity (blue) are scattered throughout, with notable concentrations around 80-100 ft and 220-240 ft. Lower resistivity can indicate a higher corrosion risk for reinforcing steel.
        \item \textbf{Cover Depth Measurements (Fig. \ref{fig6}):} The cover depth varies across the bridge, with some areas showing inadequate cover (blue areas). This is particularly evident in the first 50 ft and around the 220-250 ft mark. Insufficient cover can lead to accelerated corrosion of reinforcing steel.
    \end{enumerate}
\textbf{Overall Bridge Condition and Repair Needs:}
The bridge shows signs of deterioration and potential issues in several areas. The most concerning regions that likely need repair or further investigation are:
    \begin{enumerate}
        \item \textbf{The section from 0-50 ft:} Shows low cover depth and higher GPR attenuation.
        \item \textbf{The area around 80-100 ft:} Exhibits potential delamination, reduced concrete quality, and higher corrosion risk.
        \item \textbf{The section from 220-250 ft:} Demonstrates issues across multiple NDE techniques, including potential delamination, reduced concrete quality, higher corrosion risk, and inadequate cover depth.
    \end{enumerate}
\textbf{Recommended actions:}
    \begin{enumerate}
        \item Conduct detailed visual inspection and possibly some localized destructive testing in the identified problematic areas.
        \item Plan for concrete repairs, possibly including patching, crack injection, and localized removal and replacement of deteriorated concrete.
        \item Consider applying a protective overlay to address cover depth issues and provide additional protection against moisture ingress and chloride penetration.
        \item Implement a corrosion mitigation strategy, particularly in areas with low resistivity.
        \item Develop a regular monitoring and maintenance plan to track the progression of detected issues.
    \end{enumerate}
While the bridge doesn't appear to be in imminent danger of failure, these issues should be addressed to prevent further deterioration and ensure the long-term durability and safety of the structure.
\\
\\ \hline
\multicolumn{1}{l}{*All five images were given as an input to generate output.}
\end{tabular}
\label{tab4}
\end{center}
\end{table}

\subsection{Summarization by ChatGPT-4}

\label{summairzation}
\begin{table}[H]
\caption{Summarization by ChatGPT-4\cite{b15}}
\begin{center}
\tiny
\begin{tabular}{|p{8cm}|} 
\hline
\\
\textbf{Bridge Condition Summary and Recommendations} \\
\textbf{Overview} \\
As a structural engineer, I have analyzed the results from five different non-destructive evaluation (NDE) technologies used on a bridge. Each technology has provided insights into various aspects of the bridge’s condition, which are summarized below. \\
Findings by Image \\
\textbf{Image 0: Ground Penetrating Radar (GPR) Measurements }\\
    \begin{itemize}
        \item \textbf{Title:} Cover Depth (inch) based on Ground Penetrating Radar (GPR) Measurements.
        \item \textbf{Observation:} The cover depth is generally within the expected range, suggesting structural soundness. However, there are some areas with shallower cover.
        \item \textbf{Analysis:}
        \begin{itemize}
            \item \textbf{Good Condition:} Most of the bridge has adequate cover depth, indicating good protection for the reinforcing steel.
            \item \textbf{Areas of Concern:} Some regions with shallower cover might be vulnerable to corrosion.
        \end{itemize}
        \item \textbf{Recommendations:} Further inspection in these areas to confirm the adequacy of the cover and take necessary protective measures if required.
    \end{itemize}
\textbf{Image 1: Depth Corrected Attenuation At Top Rebar Level (dB)}
    \begin{itemize}
        \item \textbf{Observation:} The attenuation values are mostly consistent, with some higher values indicating potential issues.
        \item \textbf{Analysis:}
        \begin{itemize}
            \item \textbf{Good Condition:} Generally low attenuation values suggest minimal moisture or voids in the concrete.
            \item \textbf{Areas of Concern:} Higher attenuation areas, especially around 50-80 feet and 160-200 feet, might indicate higher moisture content or deteriorated concrete.
        \end{itemize}
        \item \textbf{Recommendations:} Conduct a detailed inspection and potentially repair or reinforce these areas.
    \end{itemize}
\textbf{Image 2: Electrical Resistivity Measurements}
    \begin{itemize}
        \item \textbf{Observation:} The red areas on the graph indicate regions of high electrical resistivity, which could suggest the presence of defects or anomalies in the material.
        \item \textbf{Analysis:}
        \begin{itemize}
            \item \textbf{Good Condition:} Blue/green areas indicate higher resistivity, suggesting less moisture and fewer conductive materials.
            \item \textbf{Areas of Concern:} Red/yellow areas suggest potential defects, such as cracks or voids.
        \end{itemize}
        \item \textbf{Recommendations:} Inspect and repair areas with low resistivity, particularly between 0-20 feet, 50-80 feet, 110-140 feet, and 200-230 feet.
    \end{itemize}
\textbf{Image 3: Frequency (kHz) based on Impact-echo (IE) Measurements}
    \begin{itemize}
        \item \textbf{Observation:} The frequency distribution shows some irregularities, especially in regions around 50-80 feet and 160-200 feet.
        \item \textbf{Analysis:}
        \begin{itemize}
            \item \textbf{Good Condition:} Most areas exhibit normal frequency values.
            \item \textbf{Areas of Concern:} Irregular frequencies may indicate delaminations or voids within the concrete.
        \end{itemize}
        \item \textbf{Recommendations:} Further investigate these regions and carry out necessary repairs to maintain structural integrity.
    \end{itemize}
\textbf{Image 4: Modulus (ksi) based on Ultrasonic Surface Waves (USW) Measurements}
    \begin{itemize}
        \item \textbf{Observation:} Variations in modulus values with lower values around 50-80 feet and 160-200 feet.
        \item \textbf{Analysis:}
        \begin{itemize}
            \item \textbf{Good Condition:} Most areas have consistent modulus values.
            \item \textbf{Areas of Concern:} Lower modulus values suggest weaker concrete in certain regions.
        \end{itemize}
        \item \textbf{Recommendations:} Reinforce or repair areas with low modulus values to ensure the bridge’s strength.
    \end{itemize}
\textbf{Overall Bridge Condition and Recommendations} \\
\textbf{Condition Summary} \\
The bridge shows potential structural issues in specific regions, particularly around:
    \begin{itemize}
        \item \textbf{50-80 feet:} Shallow cover depth, higher attenuation, irregular frequencies, and lower modulus values.
        \item \textbf{160-200 feet:} Similar issues as above, indicating potential structural weaknesses.
        \item \textbf{0-20 feet, 110-140 feet, 200-230 feet:} Lower resistivity suggesting potential defects.
    \end{itemize}
\textbf{Recommendations}
    \begin{enumerate}
        \item \textbf{Detailed Inspection:} Conduct targeted inspections in the identified regions to confirm the extent of the issues.
        \item \textbf{Repairs:}
        \begin{itemize}
            \item Increase concrete cover in areas with shallow depth.
            \item Address moisture intrusion and fill voids or delaminations.
            \item Reinforce areas with low modulus values to strengthen the structure.
        \end{itemize}
        \item \textbf{Preventative Measures:} Apply protective overlays and implement corrosion mitigation strategies.
        \item \textbf{Regular Monitoring:} Develop a monitoring plan to track the progression of detected issues and ensure timely maintenance.
    \end{enumerate}

By addressing these areas proactively, the overall condition of the bridge can be improved, potentially extending its service life and ensuring safety for its users.
\\
\\ \hline
\end{tabular}
\label{tab5}
\end{center}
\end{table}

\end{document}